\pdfoutput=1

\documentclass[11pt]{article}

\usepackage[final]{acl}

\usepackage{times}
\usepackage{latexsym}

\usepackage[T1]{fontenc}

\usepackage[utf8]{inputenc}

\usepackage{microtype}

\usepackage{inconsolata}

\usepackage{graphicx}

\usepackage{amsmath} 
\usepackage{adjustbox}
\usepackage{multirow}
\usepackage{graphicx}
\usepackage{wrapfig}
\usepackage{booktabs}
\usepackage{algorithm}
\usepackage{algpseudocode}
\usepackage{xspace}
\usepackage{wrapfig}

\newcommand{\dataset}{\texttt{LaMP-QA}\xspace}
\newcommand{\ourmethodshort}{\texttt{PlanPers}\xspace}
\newcommand{\ourmethod}{{Plan-RAG Personalization}\xspace}

\usepackage{xcolor}         
\usepackage{enumitem}
\usepackage{listings}

\title{\dataset: A Benchmark for Personalized Long-form Question Answering}

\author{Alireza Salemi \and Hamed Zamani \\
  Center for Intelligent Information Retrieval \\
  University of Massachusetts Amherst \\
  \texttt{\{asalemi,zamani\}@cs.umass.edu} \\ \\
  Data: \url{https://hf.co/datasets/alireza7/LaMP-QA} \\
  Code: \url{https://github.com/LaMP-Benchmark/LaMP-QA} \\
}

\begin{document}
\maketitle
\begin{abstract}
Personalization is essential for question answering systems that are user-centric. Despite its importance, personalization in answer generation has been relatively underexplored. This is mainly due to lack of resources for training and evaluating personalized question answering systems. We address this gap by introducing \dataset---a benchmark designed for evaluating personalized long-form answer generation. The benchmark covers questions from three major categories: (1) Arts \& Entertainment, (2) Lifestyle \& Personal Development, and (3) Society \& Culture, encompassing over 45 subcategories in total. To assess the quality and potential impact of the \dataset benchmark for personalized question answering, we conduct comprehensive human and automatic evaluations, to compare multiple evaluation strategies for evaluating generated personalized responses and measure their alignment with human preferences. Furthermore, we benchmark a number of non-personalized and personalized approaches based on open-source and proprietary large language models. Our results show that incorporating the personalized context provided leads to up to 39\% performance improvements. The benchmark is publicly released to support future research in this area.
\end{abstract}

\section{Introduction}

Personalization plays a key role in many applications such as search \cite{10.1145/1462198.1462203}, recommendation \cite{naumov2019deep}, and generation \cite{lamp, salemi2025reasoningenhancedselftraininglongformpersonalized, xu2025personalizedgenerationlargemodel}, as it contributes to improving user satisfaction and trust in the system. For information seeking, personalization is valuable as it enables systems to generate responses that are tailored to the intent, background, and preferences of the user, producing more accurate and user-specific responses. Research on personalized information seeking has predominantly focused on personalized retrieval \cite{se-pqa, 10.1162/dint_a_00104, yandex-personalized-web-search-challenge}, while the \textit{generation} aspect of the problem remains underexplored. This gap has become increasingly important with the advent of large language models (LLMs) that interact with users through natural language, making personalized text generation a critical area of study. Recently, researchers developed LaMP \cite{lamp} and LongLaMP \cite{longlamp} benchmarks for personalized text generation, however, they  exclusively focus on content generation tasks, such as email generation based on user's writing style, overlooking \textit{information seeking tasks}. This leaves a critical gap in evaluating how well LLMs can generate tailored responses to users' information needs.

A potential approach for constructing a personalized question answering dataset is human annotation, e.g., where a user poses a question and selects their preferred response from a set of generated responses. This faces two major limitations. First, the user’s selected response represents only a single sample from a broader distribution of potentially suitable responses for the user. Since the user does not have access to the full space of all possible responses, their choice may not reflect their true preference. Second, previous work has shown that effective personalization is based on access to the user’s historical interactions with the system \cite{se-pqa, peft-rag-personalization}. Collecting such history in a single annotation session is challenging, making this method difficult to scale for large and realistic datasets. A more scalable alternative is to use data from forums or community question answering (CQA) platforms. These platforms often feature questions accompanied by detailed post descriptions---serving as the \textit{question narrative}---where users explicitly articulate their specific information needs. Users often ask multiple questions over time, enabling the creation of histories that support personalized modeling and provide rich, time-based interaction data. In addition, other users in the community can respond to the question, and the question asker has the option to select a preferred ``accepted'' answer. However, relying on a single selected answer for evaluation still has limitations, as users may not see the full range of suitable responses.

An alternative evaluation strategy is to use the question narrative that accompanies the question on these platforms as a personalized question narrative. These question narratives include rich contextual cues, such as the motivation behind the question, the specific aspects the user expects to see addressed, and their primary concerns. Instead of asking users to pick a single preferred response, the question narrative can serve as a personalized evaluation rubric, defining what makes an answer satisfactory. This allows for a more principled, fine-grained evaluation based on alignment with user-specified criteria. This approach goes beyond binary or ordinal preferences, enabling systematic scoring of responses across multiple personalized criteria. This evaluation approach is also grounded in long-standing research on dataset creation in the TREC community \cite{Balog2010OverviewOT, DBLP:conf/trec/ShenCCLC07,Allan2003HARDTO, lawrie2024overviewtrec2023neuclir,Buckley1998SMARTHP}. In many TREC tasks, annotators are provided not only with the query but also with a detailed description or narrative that clarifies the underlying intent and contextual nuances of the information need. This additional information ensures more accurate and fair evaluation by helping annotators judge relevance based on the full scope of the user’s expectations. Our proposed evaluation method extends this idea to the generation setting with a focus on \textit{personalized narratives}.

This approach results in the \underline{La}nguage \underline{M}odel \underline{P}ersonalized \underline{Q}uestion \underline{A}nswering benchmark (\dataset) for training and evaluating long-form personalized answer generation systems. To collect our dataset, we begin with the SE-PQA \cite{se-pqa} dataset, designed for personalized retrieval and extracted from the StackExchange website. We filter out questions that do not require personalization, assessed by a capable LLM, Gemini \cite{gemini}, to make sure usefulness for studying personalization. Next, for the remaining questions, we filter out those where the corresponding question narrative (i.e., the post detail) do not provide details regarding the specific information needs of the user, using the same LLM. This step is crucial, as responses are evaluated based on how well they address the user's specific information needs outlined in narratives. Since question narratives are hidden during generation and used only for evaluation, we filter out questions lacking sufficient detail for a meaningful assessment. The outputs of these steps are sampled and then quality-checked by human annotators to ensure high quality. 

Inspired by prior work on personalized search \citep{se-pqa} and generation \citep{lamp}, which leverage a user’s historical interactions as the profile, we construct our benchmark by treating the user’s current question as the input, their previously asked questions as the user profile, and the key aspects extracted from the question narrative as the evaluation criteria. To evaluate a generated response, we use an LLM to assess how well the response addresses each personalized rubric aspect aspect (Examples in Figure~\ref{fig:example-dataset} and Figure~\ref{fig:example-dataset-2} in Appendix~\ref{app:dataset-creation}). This enables us to evaluate how well the response aligns with the user's needs and preferences. We divide the dataset into three categories: {(1) Arts \& Entertainment}, {(2) Lifestyle \& Personal Development}, and {(3) Society \& Culture}, with over 45 subcategories as shown in Figure~\ref{fig:dataset-dist} in Appendix~\ref{app:dataset-creation}. Examples of our benchmark are provided in Figure~\ref{fig:example-dataset} and Figure~\ref{fig:example-dataset-2} in Appendix~\ref{app:dataset-creation}, while the statistics are detailed in Table~\ref{tab:dataset-stats}.

To assess the benchmark's quality for personalization, we run a series of experiments. First, we assess the quality of automatically extracted rubric aspects---used as evaluation rubrics---from the user’s question narratives. Human annotators assign an average rating of 4.9 out of 5 on the quality of the extracted aspects, demonstrating the high quality of the aspect extraction process. We compare our proposed personalized, aspect-based evaluation method against two alternative strategies: pairwise comparison and aspect-free scoring (both also based on the question narrative). Our method achieves the highest alignment with human judgment, validating its effectiveness as an evaluation approach. Moreover, an essential characteristic of a high-quality personalized QA dataset is that the included user profile should yield better performance when used with the corresponding user's query compared to either (1) no personalization, or (2) using a profile from another user. Our results confirm this: using the target user's profile improves performance by up to 39\% over the non-personalized setting, and by up to 62\% compared to using a mismatched profile. These findings show that both the profiles and evaluation rubrics are user-specific and effectively support personalization. To establish baselines, we evaluate a range of open-source and proprietary LLMs---including Gemma 2 \cite{gemma2}, Qwen 2.5 \cite{qwen2.5}, and GPT-4o \cite{gpt4o}---in both personalized (using RAG) and non-personalized settings on the \dataset benchmark. All data and code is publicly released to encourage further research in personalized question answering.\footnote{Code is available at: \url{https://github.com/LaMP-Benchmark/LaMP-QA} and the data is available at: \url{https://hf.co/datasets/alireza7/LaMP-QA}}


\section{Problem Formulation}
\label{sec:problem-formulation}

We consider a scenario where a user \( u \) poses a question \( x_u \). Following prior work that incorporates long-term user history as a user profile \citep{se-pqa, lamp}, we assume \( P_u = \{p_i\}_{i=1}^{n_u} \), the user profile, consists of \( n_u \) user's previously asked questions along with the detailed descriptions of the question written by the user. This information facilitates a better understanding of user preferences. The objective is to use this personalized information alongside the question \( x_u \) to generate a personalized response using a QA model \( M \), expressed as \( \hat{y}_u = M(x_u, P_u) \). To evaluate the quality of the generated response, we assume access to a set of \( n_{x_u} \) personalized aspects \( E_{x_u} = \{e_i\}_{i=1}^{n_{x_u}} \) that the user expects to be addressed in response to the question \( x_u \). These aspects are extracted from a personalized question narrative \( r_{x_u} \) provided by the user. Importantly, these aspects are used exclusively for evaluation and are not accessible to the model during response generation. Finally, a metric \( \mu(x_u, \hat{y}_u, E_{x_u}, r_{x_u}) \) quantifies response quality based on the extent to which the expected aspects are covered. Since these aspects are explicitly derived from user-provided requirements, this evaluation framework enables an assessment of how well the generated response is personalized to the user's information needs.

\section{The \dataset Benchmark}
\label{sec:dataset-desc}

Personalized text generation has been studied in short- and long-form content generation, such as email writing, review writing, and email title generation, in benchmarks like LaMP \cite{lamp} and LongLaMP \cite{longlamp}. However, personalization in information-seeking differs fundamentally from these tasks. While personalized content generation focuses on mimicking the user’s writing style and preferences when generating text on their behalf, personalization in information-seeking is centered on tailoring the response to align with the user’s information needs and preferences. Although datasets for personalized retrieval, as a form of information-seeking, exist \cite{se-pqa, 10.1162/dint_a_00104, yandex-personalized-web-search-challenge}, to the best of our knowledge, no dataset focuses on answer generation. 

To construct the \dataset benchmark, we begin with the SE-PQA dataset \citep{se-pqa},\footnote{Open access under cc-by-sa 4.0 license.} which has already collected data from the StackExchange platform for retrieval tasks, to avoid scraping the website again. This website is a community-based question-answering platform where users post their questions. Each post consists of a title---phrased as a question---and a detailed description that clarifies the user's information needs. This structure allows for the formulation of a personalized question-answering task. Specifically, the post title serves as the user's question, while the detailed description outlines the key information necessary for generating an effective response for the user. Since these descriptions are written by the users themselves, they provide direct insight into their expectations, making them a valuable resource for evaluating how well a generated response aligns with their needs. Furthermore, a user’s previously asked questions can be leveraged to construct a user profile, capturing their information-seeking behavior and past interactions with the system.


\subsection{Data Collection}

\begin{table*}
    \centering
    \adjustbox{max width=\textwidth}{
    \begin{tabular}{l|ccc|ccc|ccc}
        \toprule
        \multirow{3}{*}{\textbf{Method}} & \multicolumn{3}{c}{\textbf{Arts \&}} & \multicolumn{3}{c}{\textbf{Lifestyle \& Personal}} & \multicolumn{3}{c}{\textbf{Society \&}} \\
        & \multicolumn{3}{c}{\textbf{Entertainment}} & \multicolumn{3}{c}{\textbf{Development}} & \multicolumn{3}{c}{\textbf{Culture}} \\
        \cmidrule{2-10}
        & train & validation & test & train & validation & test & train & validation & test \\
        \midrule
        \textbf{\#Questions (users)} & 9349 & 801 & 767 & 7370 & 892 & 989 & 7614 & 810 & 1074  \\
        \midrule
        \textbf{\#Evaluation Aspects} & $2.7\pm0.9$ & $4.7\pm1.2$ & $4.6\pm1.2$ & $3.1\pm1.0$ & $5.1\pm1.1$ &  $5.1\pm1.2$ & $2.9\pm0.9$ & $4.8\pm1.1$ & $4.8\pm1.0$ \\
        \midrule
        \textbf{Profile Size} & $106.7 \pm 127.3$ & $129.0 \pm 183.7$ &  $159.1 \pm 203.0$ & $116.6 \pm 162.0$ &  $98.2 \pm 198.6$ & $111.6 \pm 220.3$ &  $141.3 \pm 194.7$ & $110.5 \pm 210.6$ &  $115.8 \pm 203.6$ \\
        \midrule
        \textbf{Question Length} & $13.0 \pm 2.9$ & $10.6 \pm 4.0$ & $10.0 \pm 3.8$ & $13.6 \pm 3.3$ & $11.3 \pm 4.4$ & $11.6 \pm 4.6$ & $14.2 \pm 3.6$ & $12.1 \pm 4.9$ & $12.9 \pm 5.4$ \\
        \midrule
        \textbf{Narrative Length} & $113.1 \pm 98.2$ & $166.1 \pm 167.6$ &  $144.7 \pm 146.0$ & $132.2 \pm 104.1$ & $159.2 \pm 138.5$ & $169.4 \pm 145.2$ & $144.6 \pm 117.9$ & $161.6 \pm 158.2$ & $167.9 \pm 143.4$ \\
        \bottomrule
    \end{tabular}}
    \caption{Dataset statistics of the each category in the \dataset benchmark.}
    \label{tab:dataset-stats}
\end{table*}

The \dataset benchmark is created by adapting the SE-PQA dataset, originally designed for personalized retrieval tasks, to a personalized question answering task through the following steps:

\paragraph{Filtering out factoid questions that do not require personalization:}

Since our objective is to construct a dataset for personalized question answering, we exclude questions that are purely factual---those whose answers remain unchanged regardless of the individual asking them. In other words, we exclude factoid questions that do benefit from personalized user information. To achieve this, we use a capable LLM, employing the prompt illustrated in Figure~\ref{fig:check-personalizable} in Appendix~\ref{app:dataset-creation}. This prompt instructs the LLM to identify and label factoid questions that do not require personalization. For the test and validation sets, to ensure high-quality outputs, we utilize Gemini 1.5 Pro\footnote{Available at: \url{https://ai.google.dev/gemini-api/docs/models/gemini\#gemini-1.5-pro}} \citep{gemini} as the LLM. For the training set, we use Gemma 2 \cite{gemma2} with 27 billion parameters to reduce computational costs while maintaining strong effectiveness. To validate the effectiveness of this filtering, we manually review a sample of 500 questions from the remaining questions. This quality check ensures that the remaining questions after filtering require personalized information to improve response generation. Based on our observations, almost all the questions in this sample benefit from personalization and require some personalized context to generate high-quality responses that effectively answer the question.

\paragraph{Filtering out questions lacking sufficient information in question narrative about the response requirements:}

To evaluate a response to a given question, we extract the user's information needs from the question narrative and assess how well the response addresses them. However, some question narratives (i.e., post details) lack sufficient detail for this evaluation and do not provide clear rubrics. Thus, we filter out such questions. To achieve this, we use a capable LLM with the prompt shown in Figure~\ref{fig:check-evaluable} in Appendix~\ref{app:dataset-creation}. This prompt determines whether a question narrative explicitly specifies aspects that a response should address. If it does, the model extracts these aspects, provides supporting evidence, and explains their significance. The extracted aspects correspond to the set \( E_{x_u} \) defined in Section~\ref{sec:problem-formulation}, which we use to evaluate response personalization. Since these aspects are derived directly from user-written question narrative detailing their information needs, they are inherently personalized. For the test and validation sets, to ensure high-quality outputs, we use Gemini 1.5 Pro as the LLM. For the training set, we use Gemma 2 with 27 billion parameters to reduce costs. To evaluate the quality of the extracted rubric aspects, we conduct a human evaluation study. We sample 100 questions that passed this filtering step and present them to annotators, who rate the extracted aspects on a 1–5 scale based on their alignment with the user's stated information needs in the question narratives. Each example is independently reviewed by two annotators. The inter-annotator agreement, measured using Cohen's kappa, is \( 0.87 \), indicating a high level of consistency between reviewers. The results show an average score of \( 4.9 \) out of 5 for the extracted aspects, demonstrating their strong alignment with the information needs specified in the question narratives and confirming the high quality of the extraction. The detailed guidelines used for the human evaluation are included in Appendix~\ref{app:human-eval}.

\paragraph{Forming the \dataset benchmark:}

We use the SE-PQA train, validation, and test sets, applying the filtering steps to retain only questions that are suitable for personalized evaluation. This ensures the dataset includes questions that benefit from personalization and contain aspects reflecting the user's specific information needs. For each remaining post from user \( u \), we treat the post's question as the input \( x_u \) and extract the relevant aspects, denoted as \( E_{x_u} \), from the question narrative \( r_{x_u} \). To construct the user profile \( P_u \), we gather previous posts from the same user, capturing their information-seeking behavior. For a more fine-grained evaluation, we categorize the filtered questions into three categories: \textbf{(1) Art \& Entertainment}, \textbf{(2) Lifestyle \& Personal Development}, and \textbf{(3) Society \& Culture}. Each of the main categories consists of subcategories, totaling over 45 subcategories. The distribution of subcategories is shown in Figure~\ref{fig:dataset-dist} in Appendix~\ref{app:dataset-creation}. The dataset statistics are provided in Table~\ref{tab:dataset-stats}. Two examples from our dataset, which includes the question, question narrative, and the extracted key aspects relevant to the user, is presented in Figures \ref{fig:example-dataset} and \ref{fig:example-dataset-2} in Appendix~\ref{app:dataset-creation}.

\subsection{Evaluation Metric}
\label{sec:eval-metric}

Inspired by recent advances in the evaluation of personalized text generation \citep{salemi2025experteffectiveexplainableevaluation} and aspect-based evaluation \cite{min-etal-2023-factscore, samarinas2025factualaccuracyevaluatingcoverage}, we evaluate a generated personalized response \( \hat{y}_u \) to a user query \( x_u \) using extracted aspects from the question narrative, denoted as \( E_{x_u} \). These aspects remain hidden during response generation and are only utilized for evaluation. Specifically, we evaluate the generated response \( \hat{y}_u \) for each aspect \( e \in E_{x_u} \) using an LLM. In this paper, we use the instruction-tuned Qwen 2.5 \cite{qwen2.5} model with 32 billion parameters as the LLM for evaluation unless stated otherwise. The response is rated on a scale from 0 to 2 for each aspect, following the prompt shown in Figure~\ref{fig:eval-prompt} in Appendix~\ref{app:eval-metric}. The scores are then normalized by dividing by 2. The final evaluation score for the generated response is computed as the average of the normalized scores across all aspects \( e \in E_{x_u} \).

\section{Baselines for the \dataset Benchmark}
\label{sec:method}

This section introduces both existing and newly proposed baselines on the \dataset benchmark.

\paragraph{No-Personalization.}

For this baseline, the question \( x_u \) is provided directly to the LLM \( M \) to generate a response \( \hat{y}_u = M(x_u) \), without incorporating any personalized information, using the prompt shown in Figure~\ref{fig:baseline-no-personalization} in Appendix~\ref{app:baselines}. Since this method doesn't use the user's profile during generation, the response is generic and not personalized.

\paragraph{RAG-Personalization.}

Following \citet{lamp}, we adopt RAG to incorporate personalized context from the user's profile when answering a question. Specifically, the question \( x_u \) is used as a query to retrieve the top \( k \) relevant entries from the user's profile \( P_u \) using a retrieval model \( R \). The retrieved content is then concatenated with the question and passed to the LLM \( M \) to generate a personalized response, denoted as: \( \hat{y}_{u} = M(x_u, R(x_u, P_u, k))
\), using the prompt shown in Figure~\ref{fig:baseline-rag-personalization} in Appendix~\ref{app:baselines}. This approach allows the model to condition its generation on user-specific information, enabling the model to learn about the user preferences from its history to generate a more relevant and tailored response.

\paragraph{\ourmethod (\ourmethodshort)}

This method extends the RAG-Personalization approach by introducing a planning step prior to response generation. Given the question \( x_u \) and the top \( k \) retrieved items from the user's profile \( R(x_u, P_u, k) \), a planner model \( M_{\text{plan}} \) uses the prompt shown in Figure~\ref{fig:baseline-ours} in Appendix~\ref{app:baselines} to generate a set of aspects that are likely important to the user in the context of the question. These aspects are represented as a plan \( p_{x_u} = M_{\text{plan}}(x_u, R(x_u, P_u, k)) \). The final response is then generated by the LLM \( M \) using the prompt shown in Figure~\ref{fig:baseline-ours} in Appendix~\ref{app:baselines}, conditioned on the original question, the retrieved personalized context, and the generated plan: \( \hat{y}_{u} = M(x_u, R(x_u, P_u, k), p_{x_u}) \). This process aims to guide response generation by first explicitly inferring the aspects that the user is likely to expect in the answer, based on their question and personal history. These inferred aspects are then incorporated into the generation process, encouraging the model to produce responses that more effectively address the user's information needs.

Since the \dataset benchmark does not include reference responses, directly training the LLM \( M \) for response generation is not feasible. However, as described in Section~\ref{sec:dataset-desc}, the dataset provides a set of rubric aspects that reflect the user's expectations for the response. Leveraging this, we train a planner \( M_{\text{plan}} \) to predict these aspects based on the question \( x_u \) and the retrieved personalized context \( R(x_u, P_u, k) \). We frame this as a sequence-to-sequence \cite{seq2seq} problem and train \( M_{\text{plan}} \) using cross-entropy loss to generate the expected aspects, with each aspect title separated by a newline. During inference, the planner model predicts expected rubric aspects given the question and the personalized context. These aspects can then be used to guide the generation process, helping the model produce responses that are more aligned with the user's specific information needs.

\section{Experiments}

\begin{table*}
    \centering
    \adjustbox{max width=\textwidth}{
    \begin{tabular}{l|ccc|c}
    \toprule
    \multirow{2}{*}{\textbf{Method}} & \textbf{Arts \&} & \textbf{Lifestyle \& Personal} & \textbf{Society \&} & \textbf{Average}  \\
    & \textbf{Entertainment} & \textbf{Development} & \textbf{Culture} & \textbf{(macro)} \\
    \midrule
    \multicolumn{5}{c}{Gemma 2 Instruct (9B)} \\
    \midrule
    No-Personalization  & 0.1860 & 0.3858 & 0.4094 & 0.3270 \\
    RAG-Personalization (Random $P$) & 0.1708 & 0.3415 & 0.3310 & 0.2811 \\
    RAG-Personalization (Asker's $P_u$) & 0.2929 & 0.4232 & 0.4834 & 0.3998 \\
    \midrule
    \ourmethodshort  & \textbf{0.3548$^\dagger$} & \textbf{0.4671$^\dagger$} & \textbf{0.5481$^\dagger$} & \textbf{0.4566$^\dagger$} \\
    \midrule
    \multicolumn{5}{c}{Qwen 2.5 Instruct (7B)} \\
    \midrule
    No-Personalization  & 0.3129 & 0.4582 & 0.4769 & 0.416 \\
    RAG-Personalization (Random $P$) & 0.2547 & 0.3829 & 0.4037 & 0.3471 \\
    RAG-Personalization (Asker's $P_u$) & 0.3397 & 0.4481 & 0.4967 & 0.4281 \\
    \midrule
    \ourmethodshort  & \textbf{0.3518$^\dagger$} & \textbf{0.4818$^\dagger$} & \textbf{0.5240$^\dagger$} & \textbf{0.4525$^\dagger$} \\
    \midrule
    \multicolumn{5}{c}{GPT 4o-mini} \\
    \midrule
    No-Personalization  & 0.3713 & 0.5112 & 0.5218 & 0.4681 \\
    RAG-Personalization (Random $P$) & 0.2881 & 0.4148 & 0.4202 & 0.3743 \\
    RAG-Personalization (Asker's $P_u$) & 0.3931 & 0.4884 & 0.5310 & 0.4708 \\
    \midrule
    \ourmethodshort  & \textbf{0.4490$^\dagger$} & \textbf{0.5442$^\dagger$} & \textbf{0.6084$^\dagger$} & \textbf{0.5338$^\dagger$} \\
    \bottomrule
    \end{tabular}}
    \caption{Performance on the test set. $^\dagger$ shows a statistically significant difference between the best-performing baseline and the others using t-test ($p < 0.05$). The results on the validation set are reported in Table~\ref{tab:main-results-dev} in Appendix~\ref{app:dev-perf}.}
    \label{tab:main-results-test}
\end{table*}

\subsection{Experimental Setup}

We use a combination of open and proprietary models for the generator LLM \( M \). We employ instruction-tuned Gemma 2 (9B parameters) \cite{gemma2}, instruction-tuned Qwen 2.5 (7B parameters) \cite{qwen2.5}, and GPT-4o-mini \cite{gpt4o} as the proprietary model. For the planner model \( M_{\text{plan}} \), we use Qwen 2.5 with 7B parameters. Training details for the planner using LoRA \cite{lora} are provided in Appendix~\ref{app:setup}. All models operate with a maximum input-output token limit of 8192 and generate responses using nucleus sampling \cite{nu_sampling} with a temperature of 0.1. For retriever, we use Contriever \cite{contriever}, fine-tuned on MS MARCO \cite{msmarco}, to retrieve \( k = 10 \) items from the user profile, unless otherwise noted. Implementation details are presented in Appendix~\ref{app:setup}.

\subsection{Main Findings}
\label{sec:main-results}

\paragraph{Baselines Performance.}

The performance of the baselines on the \dataset benchmark is reported in Table~\ref{tab:main-results-test} for the test set and in Table~\ref{tab:main-results-dev} in Appendix~\ref{app:dev-perf} for the validation set. The results in Table~\ref{tab:main-results-test} demonstrate that \ourmethod significantly outperforms all baselines across all evaluation categories. This highlights the utility of the training data provided by the \dataset benchmark for training an effective planner model that can infer the key information users expect in response to their questions. Furthermore, the results show that all personalized baselines leveraging the asker's profile to tailor the LLM's response outperform the non-personalized model in nearly all cases. This indicates that incorporating user-specific context enhances the relevance and quality of the generated responses for each user, underscoring the importance of personalization in question answering. 

Moreover, we observe that the best performance is achieved when GPT-4o-mini is used as the backbone LLM for generation. Notably, the highest performance occurs when \ourmethod is applied to this model, highlighting the effectiveness of the planning step in understanding user preferences and incorporating them into the generated response. This demonstrates that even with a strong underlying LLM, explicitly modeling user intent through planning provides substantial gains in personalized question answering.

\begin{figure}    
    \centering
    \includegraphics[width=\linewidth]{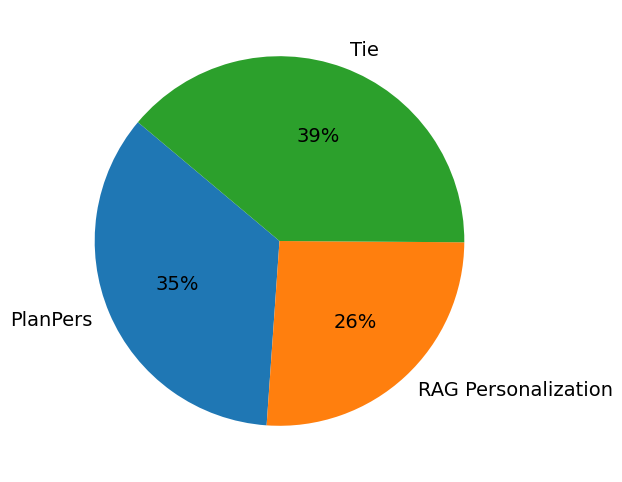}
    \caption{Results of pairwise human evaluation between RAG-Personalization and \ourmethodshort with Qwen 2.5 Instruct 7B as the backbone LLM. Each slice reports the winning percentage of the corresponding method.}
    \label{fig:human-eval}
\end{figure}

To further assess the comparative quality of generated responses, we conducted a human evaluation between the two strongest baselines, \ourmethodshort and RAG-Personalization, both with Qwen 2.5 Instruct as the backbone. We randomly sampled 100 outputs from each system and asked two human annotators to evaluate them. For each instance, annotators were provided with the question, its narrative, and the personalized rubrics, and were asked to select the better response or indicate a tie. The results, shown in Figure~\ref{fig:human-eval}, indicate that \ourmethodshort was preferred in 35\% of cases, while RAG-Personalization was preferred in 26\% of cases. The remaining instances were judged as ties. These findings suggest that \ourmethodshort produces responses that are more consistently aligned with the question narrative and personalized rubrics from a human perspective.

\begin{figure*}
    \centering
    \includegraphics[width=\textwidth]{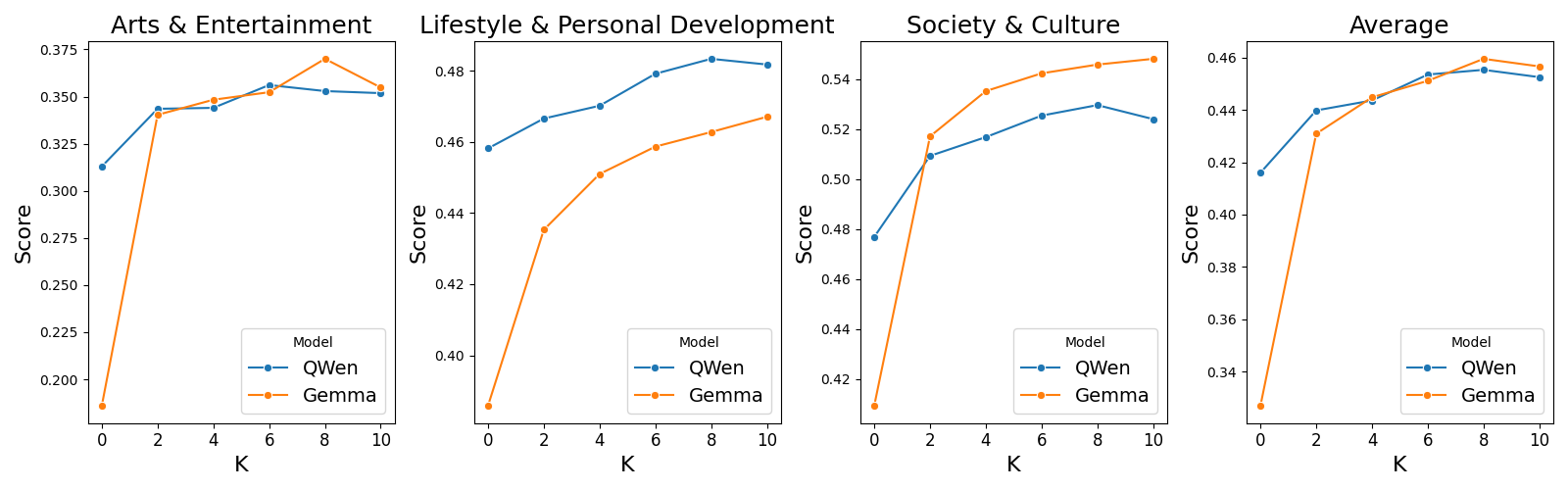}
    \vspace{-0.8cm}
    \caption{Effect of number of items ($K$) from the user's profile on the performance of \ourmethod on the test set. The results for the validation set is shown in Figure~\ref{fig:ctx-size-perf-dev} in Appendix~\ref{app:dev-perf}.}
    \label{fig:ctx-size-perf-test}
\end{figure*}

\paragraph{Effect of Personal Data in the \dataset Benchmark on the Performance.}

A crucial criterion for a dataset designed for personalized question answering is that incorporating personalized context, such as a user’s history, into the response generation process should lead to improvements in the preferability of the generated response, as assessed by a metric grounded in the user’s information needs. To assess the \dataset benchmark under this criterion, we conduct three experiments.

In the first experiment, we compare a non-personalized LLM with a personalized LLM that uses RAG without any additional training. As shown in Table~\ref{tab:main-results-test}, incorporating personalized information consistently improves performance across almost all cases. This demonstrates that the user-specific context provided in the \dataset benchmark is effective in enabling LLMs to generate more tailored and relevant responses.

In the second experiment, we aim to assess whether the user profiles provided in the \dataset benchmark are indeed tailored to individual users and beneficial for generating personalized responses. To this end, we use the RAG approach but retrieve information from a random user profile rather than from the actual asker's profile. The results, presented in Table~\ref{tab:main-results-test} for the test set and Table~\ref{tab:main-results-dev} in Appendix~\ref{app:dev-perf} for the validation set, reveal a significant performance drop when using random profiles. In fact, this setting performs worse than the non-personalized baseline, highlighting that the retrieved context must be user-specific to be helpful. These findings confirm that the user profiles in the \dataset benchmark are aligned with the users' rubric-based expectations and are useful for studying personalized question answering.

In the third experiment, we investigate the effect of varying the amount of retrieved information from the user profile on the performance of \ourmethod, the best-performing baseline. Specifically, we vary the number of retrieved items \( k \in \{0, 2, 4, 6, 8, 10\} \) and report the model's performance on the test set in Figure~\ref{fig:ctx-size-perf-test}, and on the validation set in Figure~\ref{fig:ctx-size-perf-dev} in Appendix~\ref{app:dev-perf}. The results show improvement in performance as more items are retrieved from the user profile, suggesting that incorporating a larger amount of personalized context helps the model better infer and address the user’s preferences. This highlights the importance of rich user history in effective personalization.\footnote{Due to the 8192-token limit of Gemma 2, we restrict the number of retrieved items to a maximum of 10.}

\paragraph{Headroom Analysis for Improving Planning in \ourmethodshort.}

\begin{figure}    
    \centering
    \includegraphics[width=\linewidth]{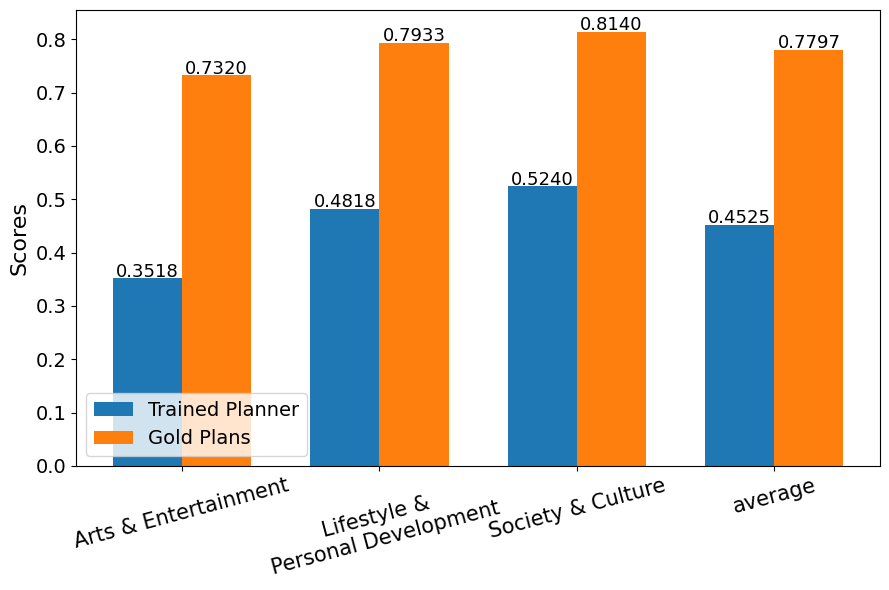}
    \vspace{-0.8cm}
    \caption{Comparison between trained planner and utilizing gold plans on the performance of \ourmethodshort (Qwen 2.5 Instruct 7B) on the test set of \dataset.}
    \vspace{-0.4cm}
    \label{fig:planner-vs-gold}
\end{figure}

To evaluate the potential upper bound of plan quality in \ourmethodshort, we conduct an experiment comparing our trained planner with an oracle setting. In the oracle condition, instead of generating plans using the learned planner, we directly use gold-standard plans derived from the rubric aspects associated with each user. These gold plans are then provided to the model to generate responses. The results of this comparison, shown in Figure~\ref{fig:planner-vs-gold} for Qwen 2.5 Instruct, indicate that access to gold plans substantially improves performance, yielding gains of 155\%–208\% on different tasks. While the trained planner already achieves significant improvements over baseline methods (Table~\ref{tab:main-results-test}), this gap highlights considerable headroom for further enhancing the planner’s ability to generate high-quality and user-specific plans.

\paragraph{Effect of Evaluation Method and Auto-Rater Size.}

One might ask about alternative methods for evaluating generated responses. We compare three approaches, each aiming to assess how well a response aligns with the user's information needs. First, inspired by prior work \cite{gemba, geval}, we use a direct scoring approach in which, given the question, the question narrative representing the user's stated information need, and the generated response, the LLM assigns a score between 0 and 1 using the prompt shown in Figure~\ref{fig:eval-prompt-without-aspects} in Appendix~\ref{app:eval-metric}. Second, we evaluate responses in a pairwise setup, where the LLM is provided with the question, the user's stated information need, and two candidate responses, and is asked to choose the better one using the prompt in Figure~\ref{fig:eval-prompt-pairwise}. Lastly, we use our proposed evaluation method in Section~\ref{sec:eval-metric}, which scores responses based on how well they address the individual aspects extracted from the user's information needs. The implementation details for all evaluation approaches are provided in Appendix~\ref{app:eval-metric}.

In the pairwise setting, we observe a strong position bias that undermines the reliability of this approach. Consistent with prior findings by \citet{salemi2025experteffectiveexplainableevaluation}, our experiments show that instruction-tuned Qwen 2.5 (32B parameters) changes its preferred response in 78\% of cases when the order of the two responses is reversed. This high sensitivity to position shows that the model's judgments are heavily influenced by presentation order, making this method unsuitable for robust and fair assessment of responses.

To assess the effectiveness of the other two methods, we present 100 examples from the test set to human annotators. Each example includes two generated responses—one from \ourmethod and one from RAG-Personalization, both using Qwen 2.5. Annotators are instructed to choose the response that better addresses the information need from the question narrative. Each example is independently evaluated by two raters. The inter-annotator agreement, measured using Cohen’s kappa, is \( 0.726 \), indicating moderate to substantial agreement and validating the reliability of the human evaluation process. The results show that in \( 73\% \) of cases, the aspect-based evaluation method selects the same response as human annotators, while the direct scoring method aligns with human preferences in only \( 58\% \) of cases. This demonstrates that the proposed aspect-based evaluation approach achieves a higher alignment with human judgments, highlighting its effectiveness as a more reliable and fine-grained evaluation strategy for personalized question answering.

\begin{figure}    
    \centering
    \includegraphics[width=\linewidth]{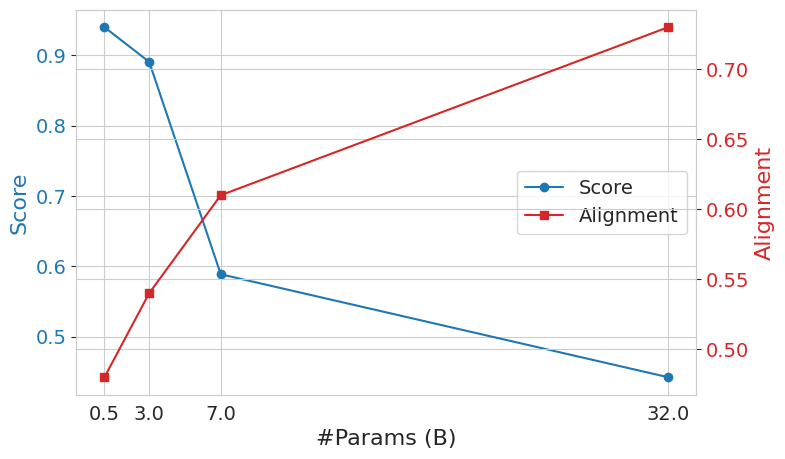}
    \vspace{-0.4cm}
    \caption{Trade-off between assigned score and alignment with human using the aspect-based evaluation metric in Section~\ref{sec:eval-metric} across different evaluator sizes.}
    \label{fig:score-param-human-alignemnet}
\end{figure}

In another experiment, we observed that the number of parameters of the evaluator LLM significantly impacts both its alignment with human judgment and the scores it assigns to the responses. We report the alignment with human preferences and the average score assigned to the generated responses using the aspect-based evaluation approach with Qwen2.5 models of varying sizes---0.5B, 3B, 7B, and 32B parameters---in Figure~\ref{fig:score-param-human-alignemnet}. The results show a clear trend: as the size of the evaluator LLM increases, its alignment with human judgment improves, while the average score it assigns to outputs decreases. Specifically, with a 0.5B parameter model, alignment with human preferences is only 48\%, and the average score assigned is relatively high at 0.94. In contrast, the 32B evaluator achieves a much higher alignment of 73\% but assigns a significantly lower average score of 0.44. This suggests that larger evaluator LLMs are better at capturing nuanced quality signals aligned with human expectations. We found that smaller LLMs are less capable of distinguishing varying degrees to which a response addresses a particular aspect. Instead, they tend to behave like binary classifiers without the ability to assess how well it is addressed. This leads to inflated scores and weaker alignment with human evaluators.


\section{Related Work}

\paragraph{Personalized LLMs.}

Personalization plays a central role in search, recommendation, and text generation \citep{10.1145/2702123.2702503, 10.1145/1462198.1462203, naumov2019deep, lamp}. To personalize an LLM, \citet{lamp} proposed a RAG framework that retrieves information from the user profile and incorporates it into the prompt provided to the LLM. Existing methods span a range of strategies, including training retrievers with relevance feedback \citep{rspg}, optimizing LLMs using user-specific supervision \citep{jang2023personalized}, and creating personalized prompts tailored to the user \citep{Li_2024}. Parameter-efficient fine-tuning have been proposed for personalized generation \citep{tan2024personalized}, with recent work integrating such methods into RAG pipelines \citep{peft-rag-personalization}. In addition, reasoning and self-training have shown effectiveness in improving long-form personalized generation \citep{salemi2025reasoningenhancedselftraininglongformpersonalized}. Personalized assistants have also been investigated in recent work \citep{li2023teach, mysore2023pearl, lu2024corporate, zhang-etal-2024-llm-based}. Despite growing interest in personalized NLP, personalized question answering remains underexplored.

\paragraph{Personalized Information Seeking.}

Personalized text generation has been explored in both short- and long-form generation settings, including tasks such as email subject line generation and social media post generation, as demonstrated in benchmarks like LaMP \cite{lamp} and LongLaMP \cite{longlamp}. However, these benchmarks focus solely on content generation and do not cover information-seeking. On the other hand, personalized information-seeking has been studied in the context of retrieval, using datasets such as SE-PQA \cite{se-pqa}, AOL4PS \cite{10.1162/dint_a_00104}, and the Personalized Web Search Challenge \cite{yandex-personalized-web-search-challenge}. With the increasing adoption of generative AI systems, it is crucial to revisit this problem from a generation perspective---an area that remains underexplored despite its importance. This paper addresses this gap by introducing a the \dataset benchmark specifically designed to evaluate personalized question answering with LLMs.

\section{Conclusion}

This paper introduced the \dataset benchmark, specifically designed to evaluate personalized question answering with LLMs. The benchmark spans three broad domains: Arts \& Entertainment, Lifestyle \& Personal Development, and Society \& Culture. We conducted a comprehensive analysis to assess the benchmark’s effectiveness in facilitating the evaluation of personalization in question answering. We investigate multiple evaluation strategies and find that aspect-based evaluation achieves the highest alignment with human judgment. We evaluate standard RAG baselines and introduce novel planning-based methods for generating personalized responses that align more with the user's information needs. Experimental results demonstrate that leveraging personalized context alongside planning-based personalized response generation leads to substantial improvements in response quality and personalization.

\section*{Acknowledgment}

We would like to thank Cheng Li, Mingyang Zhang, Qiaozhu Mei, Weize Kong, Tao Chen, Zhuowan Li, Spurthi Amba Hombaiah, and Michael Bendersky for their valuable feedback and generous support in preparing this work and shaping the formulation of the problem. This work was supported in part by the Center for Intelligent Information Retrieval, in part by NSF grant \#2143434, in part by NSF grant \#2402873, and in part by Google. Any opinions, findings and conclusions or recommendations expressed in this material are those of the authors and do not necessarily reflect those of the sponsor.

\section*{Limitations}

This work has the following limitations:

\paragraph{On the Automatic Evaluation of Personalization.}

Although the proposed evaluation method demonstrates strong alignment with human judgments, the automatic evaluation of personalized text generation remains a challenging problem \cite{lamp, salemi2025experteffectiveexplainableevaluation, salemi2025reasoningenhancedselftraininglongformpersonalized, longlamp}, as the most reliable evaluator is the original user who posed the question. However, consistently obtaining feedback from the same user across different studies is impractical, costly, and in many cases impossible. As a result, automatic evaluation methods are necessary to enable scalable and reproducible assessment of personalization quality. Note that this challenge is inherent to the personalization problem itself and is not specific to our or any personalization evaluation dataset, as also noted by previous work \cite{salemi2025experteffectiveexplainableevaluation, longlamp}. 

\paragraph{On the Completeness of Question Narratives.}

One underlying assumption in this work is that users articulate all of their information needs within the question narrative. However, this is not always the case—users may be uncertain about what they expect in a response or may omit relevant aspects unknowingly. To address this, and following prior work on nugget-based evaluation of response generation \cite{pradeep2025greatnuggetrecallautomating}, we adopt a recall-oriented evaluation strategy. Under this approach, a response is considered satisfactory if it successfully addresses all explicitly stated user aspects, without penalizing the model for not covering unstated or implicit needs. Capturing the information needs that users are unaware of remains a difficult and often infeasible task. Nonetheless, developing techniques to handle such cases represents an important future direction for more comprehensive evaluation of personalization in question answering.

\paragraph{On the Privacy Considerations of Personalization.}

Personalizing LLMs necessitates access to user-specific data in order to effectively tailor responses. However, the use and sharing of personal data raises important privacy concerns that must be carefully addressed in the development of effective personalized language models \cite{li2023teach,peft-rag-personalization, 10.1145/345124.345155}. While this paper does not investigate privacy-preserving approaches, it focuses solely on enabling research by providing publicly available data for studying personalized question answering. Addressing the privacy implications of personalized LLMs and resolving them remains an important future work.

\paragraph{On the Model Size and Families.}

There exist various families of open-source and proprietary large language models (LLMs) with different model sizes. However, due to computational and financial constraints, this paper focuses on evaluating two widely used open-source model families, Gemma \cite{gemma2} and Qwen \cite{qwen2.5}, at a specific size. While a comprehensive comparison across model architectures and sizes would be valuable, it falls beyond the scope of this work. Our primary goal is not to benchmark model performance exhaustively, but rather to establish representative baselines for the proposed \dataset benchmark. Nonetheless, we acknowledge this as a limitation of our study, though not a fundamental flaw of the dataset itself.

\bibliography{custom}

\appendix

\section{Filtering Prompts \& Examples from the \dataset Benchmark}
\label{app:dataset-creation}

\begin{figure*}[!ht]
    \centering
    \includegraphics[width=\textwidth]{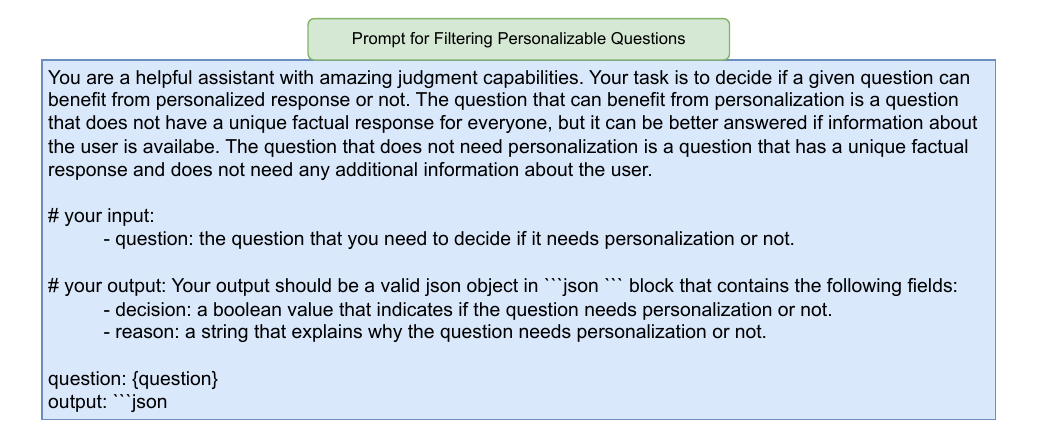}
    \caption{Prompt used with Gemini 1.5 Pro to evaluate whether the question can benefit from personalization.}
    \label{fig:check-personalizable}
\end{figure*}

\begin{figure*}[!ht]
    \centering
    \includegraphics[width=\textwidth]{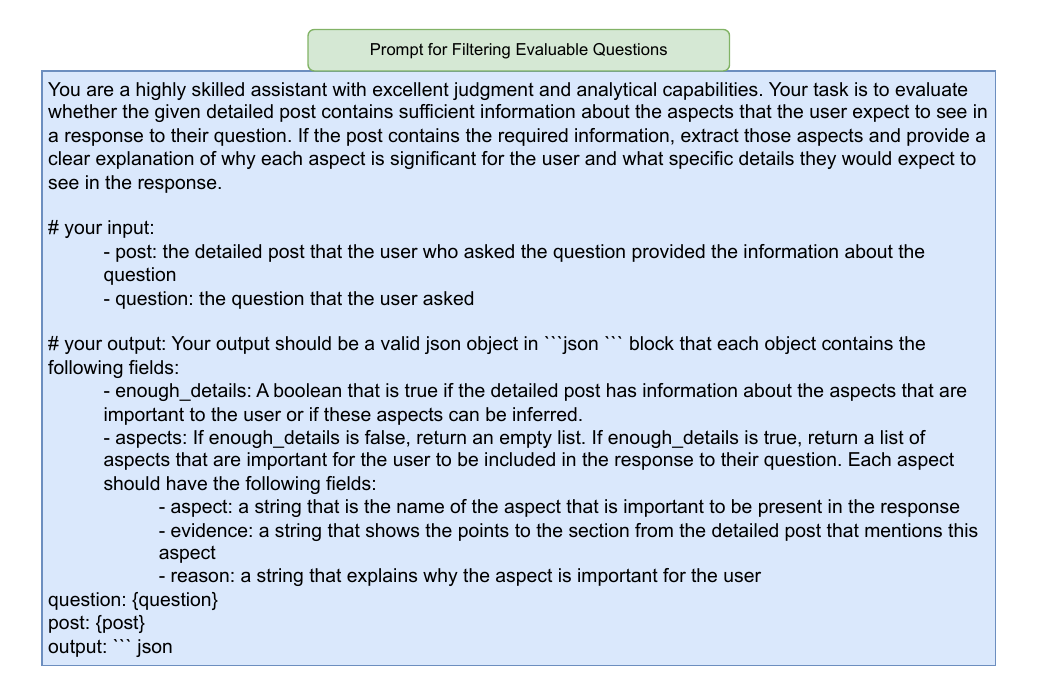}
    \caption{Prompt used with Gemini 1.5 Pro to assess whether sufficient information is available for evaluation and extract key rubric aspects from the question narrative.}
    \label{fig:check-evaluable}
\end{figure*}

The \dataset benchmark employs Gemini 1.5 Pro to refine the SE-PQA \cite{se-pqa} dataset by filtering out factoid questions that do not benefit from personalization, using the prompt provided in Figure~\ref{fig:check-personalizable}. Furthermore, the same model is used to exclude questions that lack sufficient information in the question narrative (i.e., post details) necessary for evaluating response requirements, based on the prompt illustrated in Figure~\ref{fig:check-evaluable}. 

\begin{figure*}[!ht]
    \centering
    \includegraphics[width=\textwidth]{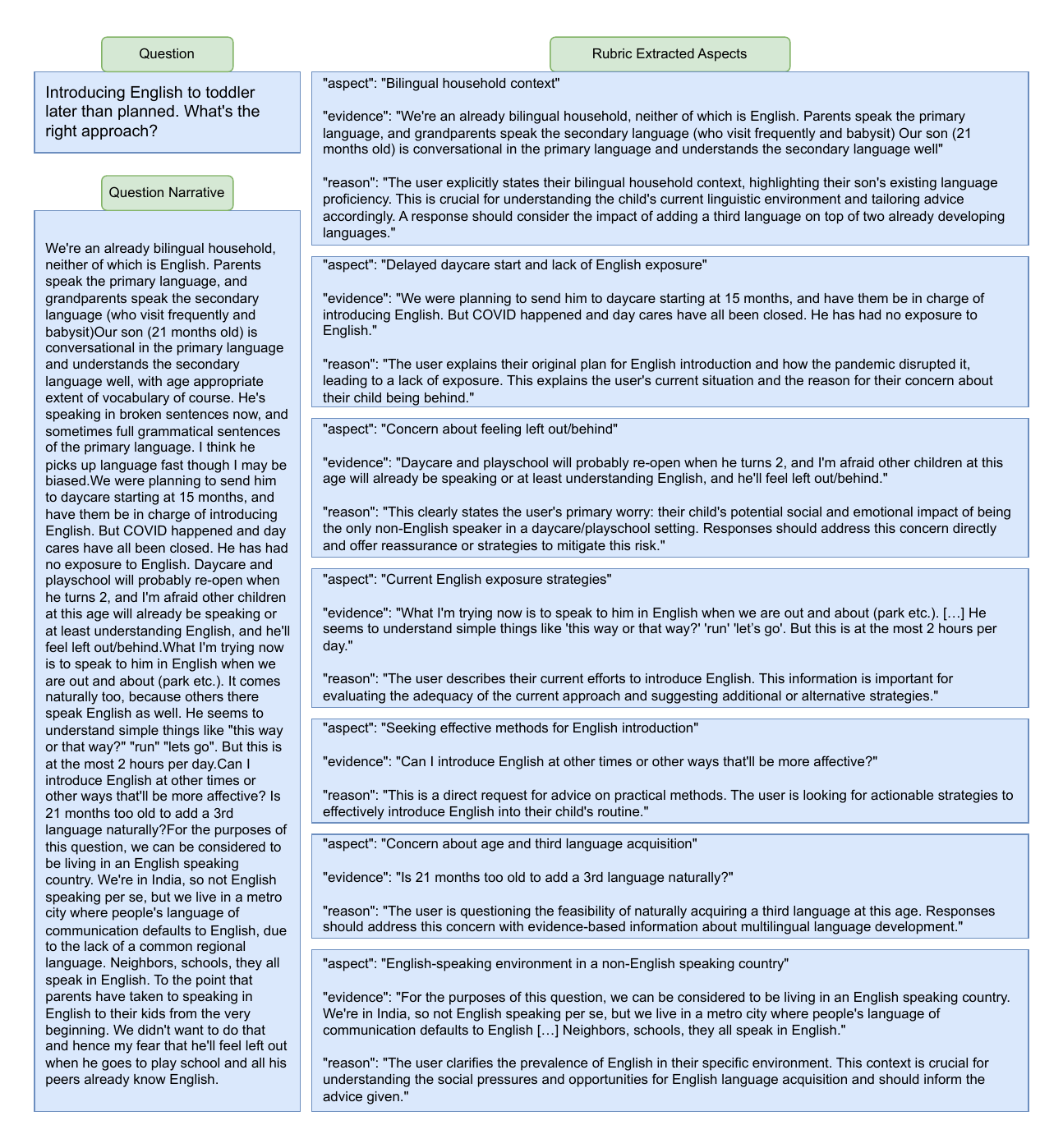}
    \caption{Example 1 of a question, the question narrative (i.e., post details), and the extracted important aspects that the user expects to be addressed in the response from the \dataset benchmark.}
    \label{fig:example-dataset}
\end{figure*}

\begin{figure*}[!ht]
    \centering
    \includegraphics[width=\textwidth]{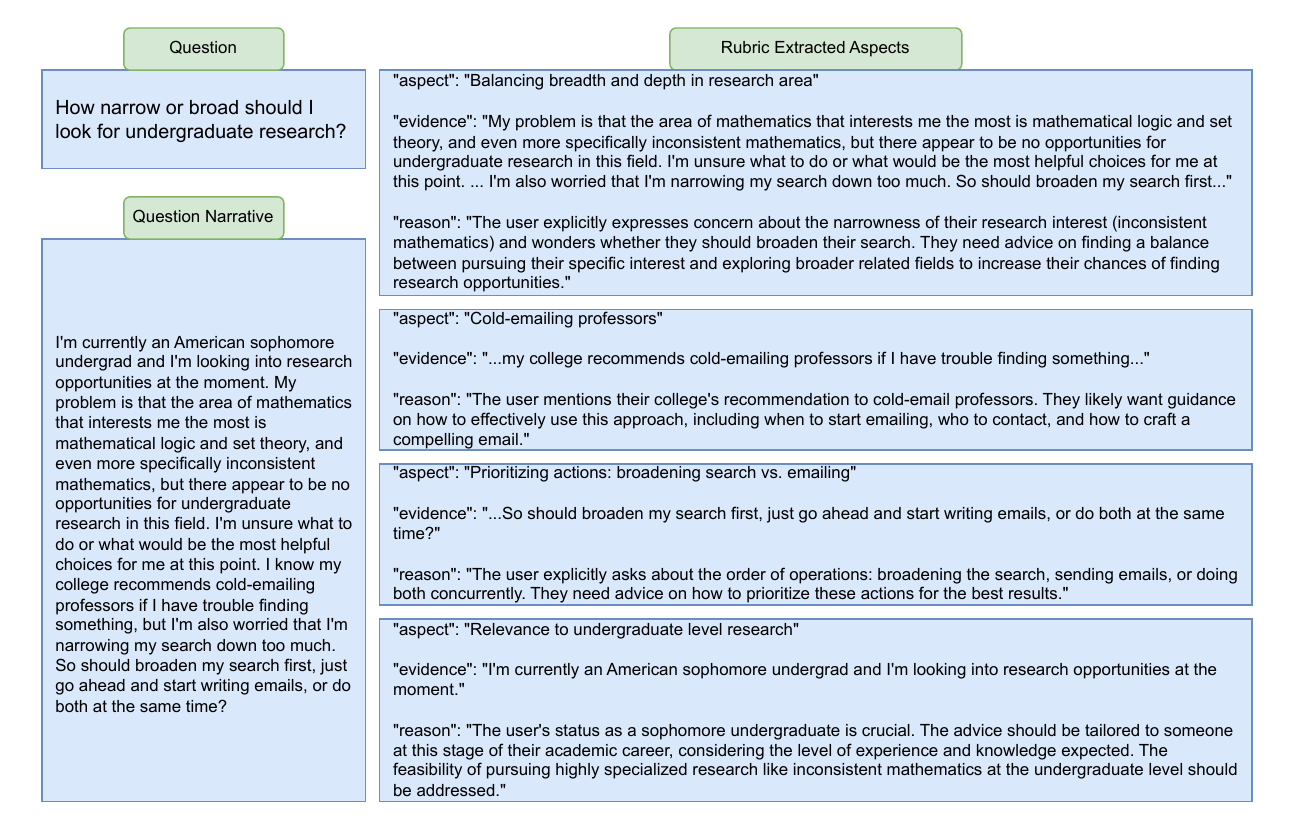}
    \caption{Example 2 of a question, the question narrative (i.e., post detail), and the extracted important aspects that the user expects to be addressed in the response from the \dataset benchmark.}
    \label{fig:example-dataset-2}
\end{figure*}

To illustrate the structure of questions, question narrative, and rubric aspects in the \dataset benchmark, Figures~\ref{fig:example-dataset} and~\ref{fig:example-dataset-2} present two representative examples from the dataset. These examples demonstrate how the benchmark captures user-specific information needs and the corresponding personalized evaluation criteria.

\begin{figure*}[!t]
    \centering
    \includegraphics[width=\textwidth]{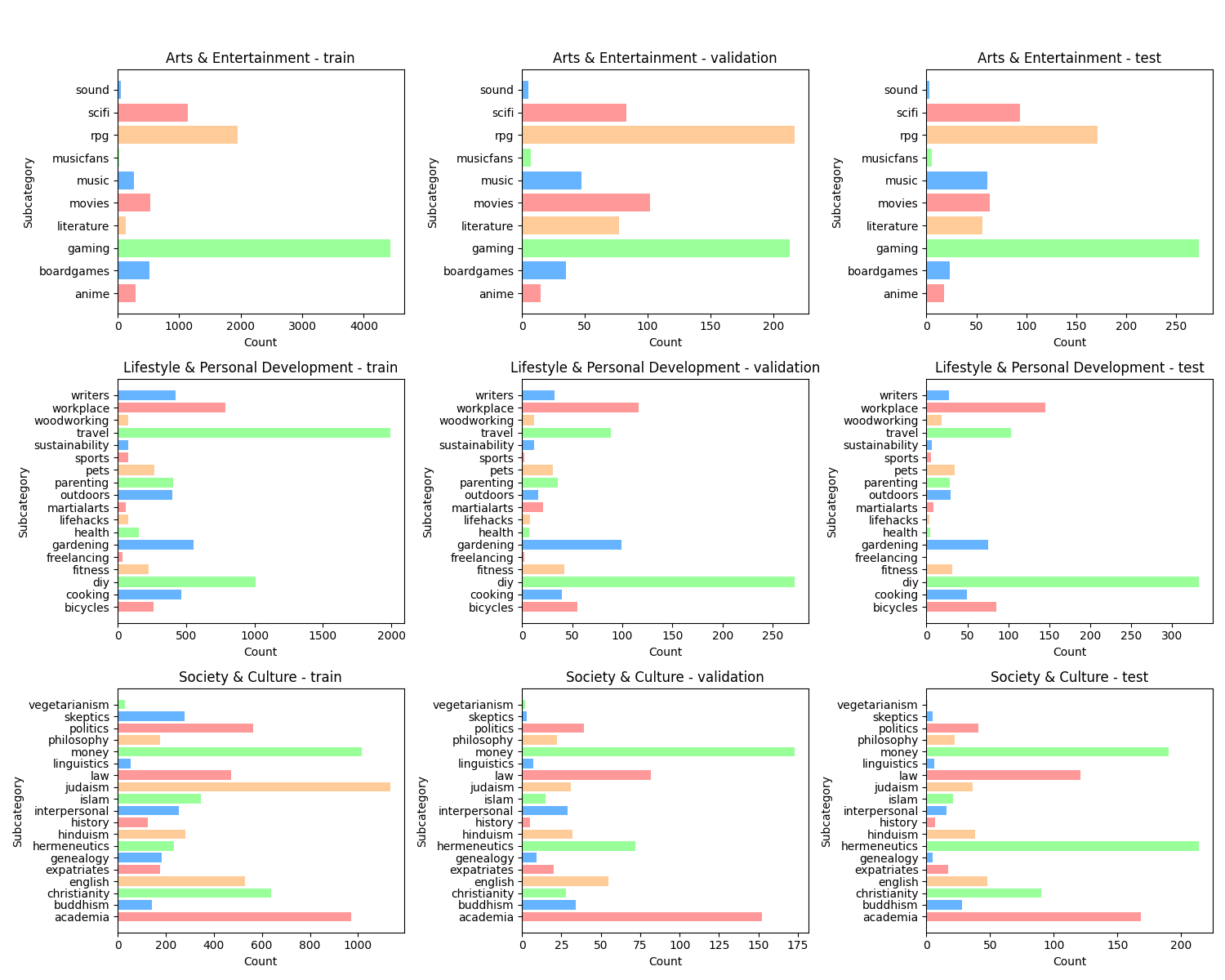}
    \vspace{-0.5cm}
    \caption{Distribution of subcategories in train, validation, and test sets of the \dataset benchmark.}
    \label{fig:dataset-dist}
\end{figure*}

\section{Human Annotation Instructions}
\label{app:human-eval}

We conduct two types of human annotation in this study. In the first experiment, we present annotators with a question, its corresponding question narrative, and the set of automatically extracted aspects obtained using Gemini 1.5 Pro as the extraction model. Annotators are instructed to evaluate the quality of the extracted aspects based on how well they reflect the user’s stated information needs, as described in the question narrative, using the following criteria:
\begin{lstlisting}[breaklines=true]
5: The generated aspects contain all the important information needs mentioned in the post detail.
4: The generated aspects contain most of the important information needs mentioned in the post detail.
3: The generated aspects contain some of the important information needs mentioned in the post detail but missed a few of them.
2: The generated aspects contain some of the important information needs mentioned in the post detail but missed some of them.
1: The generated aspects contain a few of the important information needs mentioned in the post detail and missed most of them.
\end{lstlisting}

In the second experiment, annotators are provided with two generated responses for a given question, along with the corresponding question narrative that outlines the user’s information need. Based on the information specified in the question narrative, annotators are asked to determine which of the two responses better addresses the user's stated requirements and to select the preferred response accordingly:
\begin{lstlisting}[breaklines=true]
Given a question, its corresponding post detail, and two generated responses, please evaluate which response best addresses the user's information need as described in the post detail. You may select:
    - Response 1 if it better satisfies the user's stated requirements,  
    - Response 2 if it better satisfies the user's stated requirements, or  
    - Tie if both responses are equally satisfactory in addressing the question based on the post detail.

Question: [question]
Post Detail: [details]

Response 1: [response 1]
Response 2: [response 2]
\end{lstlisting}

\section{Evaluation Metric Details}
\label{app:eval-metric}

In this paper, we explore three distinct evaluation strategies for assessing the quality of generated responses to user questions. We compare the outcomes of these methods against human preference judgments to identify the most effective evaluation approach for the \dataset benchmark.

\paragraph{LLM-Based Directly Scoring Using User-Stated Information Needs:}

\begin{figure*}[!ht]
    \centering \includegraphics[width=\textwidth]{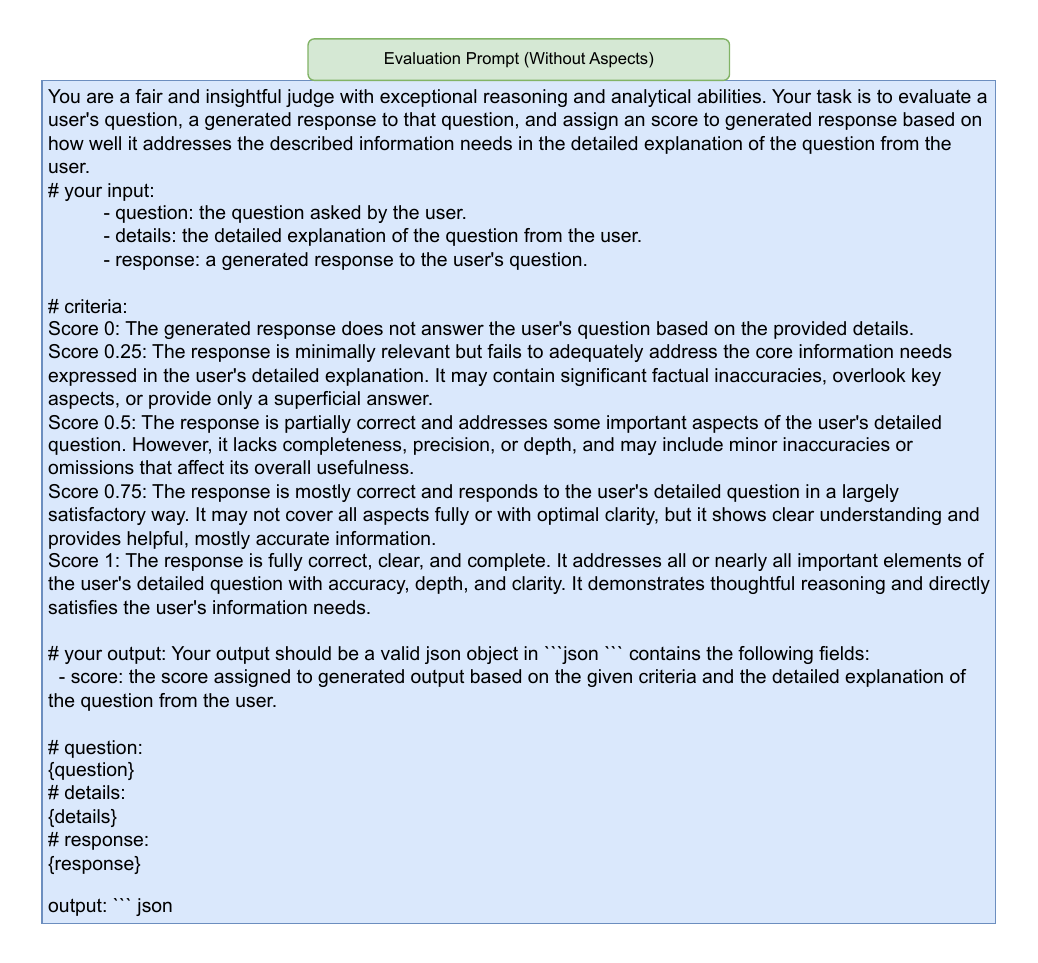}
    \caption{Evaluation prompt used for assessing the quality of generated personalized responses without the extracted aspects.}
    \label{fig:eval-prompt-without-aspects}
\end{figure*}

In this method, we provide the LLM with the question, the corresponding question narrative that specify the user's information needs, and the generated response, along with a set of evaluation criteria defined in the prompt shown in Figure~\ref{fig:eval-prompt-without-aspects}. The LLM evaluates the generated output based on how well it aligns with the stated information needs and assigns a score on a 5-point scale: {0, 0.25, 0.5, 0.75, 1}. This approach relies on the LLM's ability to interpret and reason over the user's expectations as expressed in the question narrative.

\paragraph{LLM-Based Pairwise Preference Using User-Stated Information Needs:}

In this method, the LLM is presented with the question, the corresponding question narrative that articulate the user's information needs, and two candidate responses. The model is then asked to select the response that better addresses the user's question, taking into account the specified information needs. However, consistent with prior findings by \citet{salemi2025experteffectiveexplainableevaluation}, we observe that this approach suffers from significant position bias. Specifically, using instruction-tuned Qwen 2.5 (32B parameters), we find that simply reversing the order of the responses alters the LLM’s preference in 78\% of cases. This high sensitivity to response position highlights the unreliability of this strategy for robust assessment.

\begin{figure*}[!ht]
    \centering \includegraphics[width=\textwidth]{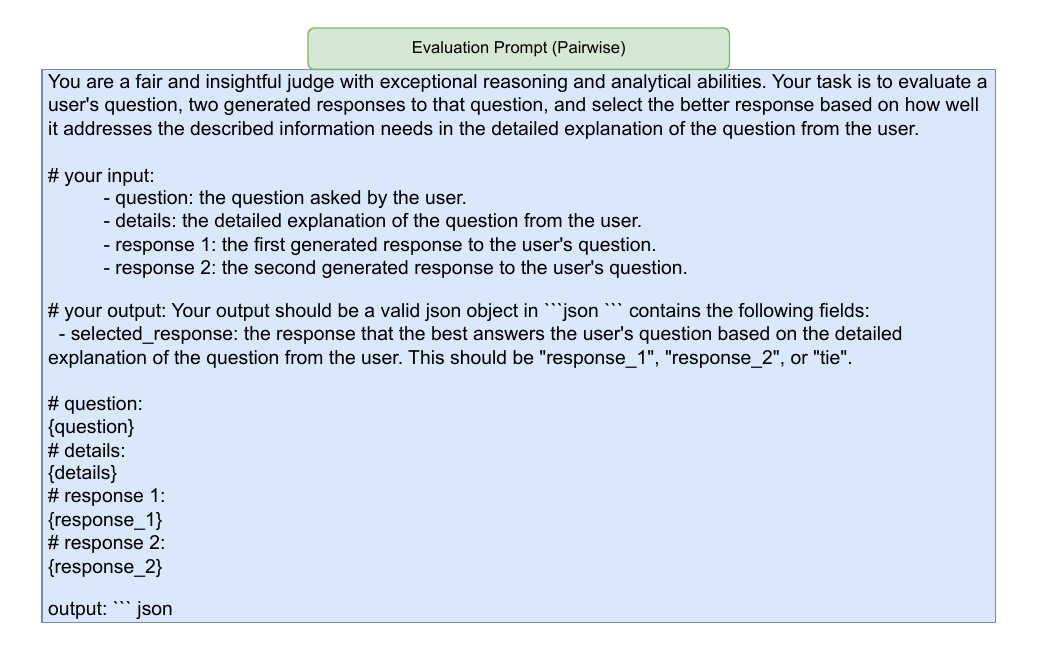}
    \caption{Evaluation prompt used for assessing the quality of generated personalized responses with pairwise evaluation.}
    \label{fig:eval-prompt-pairwise}
\end{figure*}

\paragraph{Aspect-Based Evaluation Using User Information Needs:}

\begin{figure*}[!ht]
    \centering \includegraphics[width=\textwidth]{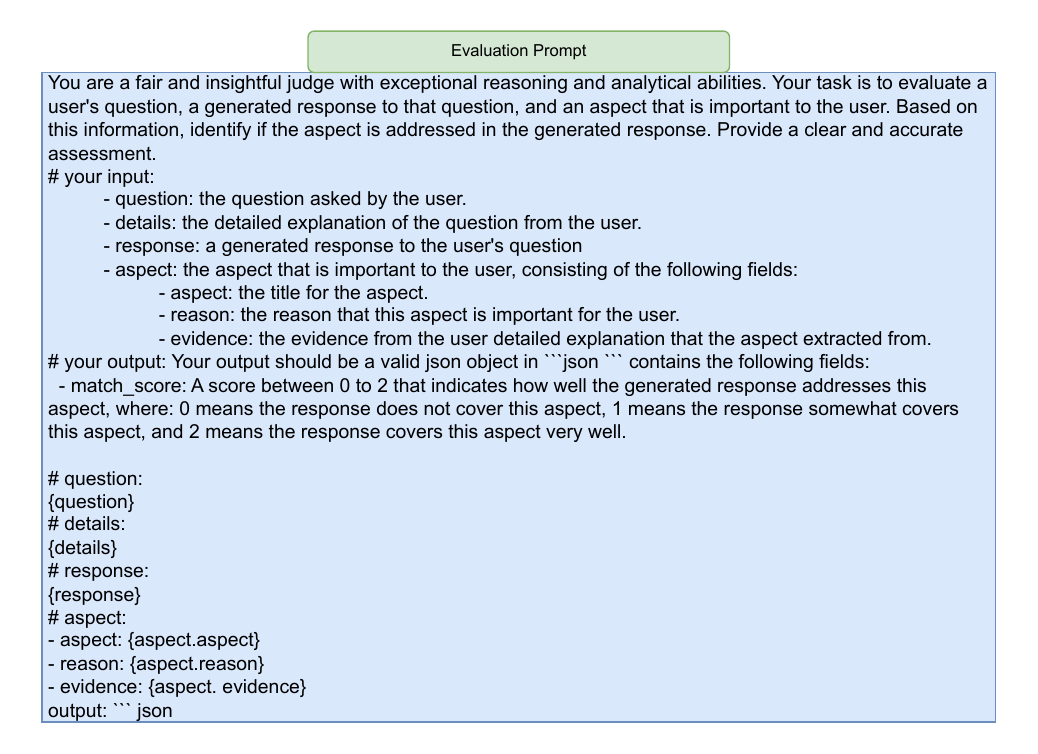}
    \caption{Evaluation prompt used for assessing the quality of generated personalized responses using the extracted aspects.}
    \label{fig:eval-prompt}
\end{figure*}

This evaluation strategy, described in Section~\ref{sec:eval-metric}, leverages the set of extracted important aspects derived from the user's stated information needs. For each aspect, the LLM assesses how well the generated response addresses it, assigning a score between 0 and 2, which is then normalized by dividing by 2. The prompt used for this evaluation is shown in Figure~\ref{fig:eval-prompt}. The final score for the response is computed as the average of the normalized scores across all aspects. 
The method is formally defined in Algorithm~\ref{alg:eval}. The key distinction between this method and direct scoring using an LLM is that it evaluates each aspect individually, assigning a score to each user-relevant aspect rather than providing a single overall score for the entire response. This aspect-level evaluation enables a fine-grained assessment of how well the response aligns with the user's information needs.

\begin{algorithm*}
\caption{Implementation of evaluation metric for the \dataset benchmark.}\label{alg:eval}
\begin{algorithmic}[1]
\Require prompt $x_u$, requirement details $r_u$, important aspects $E_{x_u}$, generated response $\hat{y}_{x_u}$, Evaluator LLM $\pi$
\Ensure score $s_{\hat{y}_{x_u}}$
\State $s^{t}_{\hat{y}_{x_u}} = 0$ \Comment{Score initialization with zero}
\For{$e_i \in E_{x_u}$} \Comment{Scoring output based on each aspect}
\State $s^{t}_{e_i} = \pi(x_u, r_u, e_i, \hat{y}_{x_u})$ \Comment{Scoring output based on aspect using prompt in Figure~\ref{fig:eval-prompt}}
\State $s_{e_i} = \frac{s^{t}_{e_i}}{2}$ \Comment{Normalizing the aspect score for aspect by division by 2}
\State $s^{t}_{\hat{y}_{x_u}} = s^{t}_{\hat{y}_{x_u}} + s_{e_i}$ \Comment{Score accumulation for averaging}
\EndFor
\State $s_{\hat{y}_{x_u}} = \frac{s^{t}_{\hat{y}_{x_u}}}{|E_{x_u}|}$ \Comment{Averaging the output score using division by the number of aspects}
\State \Return $s_{\hat{y}_{x_u}}$ \Comment{Returning score for output for user $u$}
\end{algorithmic}
\end{algorithm*}

\section{Prompt Templates}
\label{app:baselines}

\begin{figure*}[!ht]
    \centering \includegraphics[width=\textwidth]{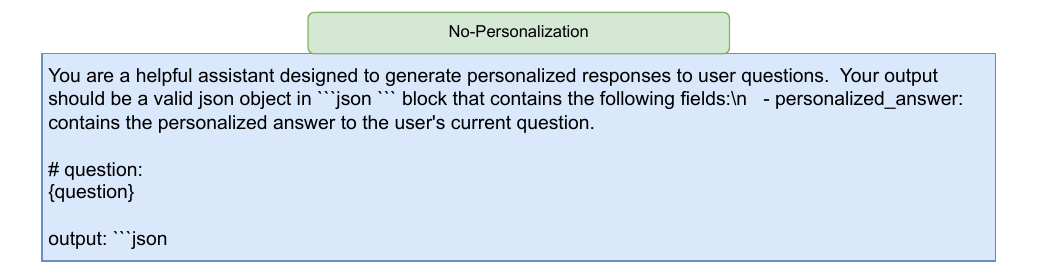}
    \caption{Prompt used with LLMs that do not incorporate personalization.}
    \label{fig:baseline-no-personalization}
\end{figure*}

\begin{figure*}[!ht]
    \centering \includegraphics[width=\textwidth]{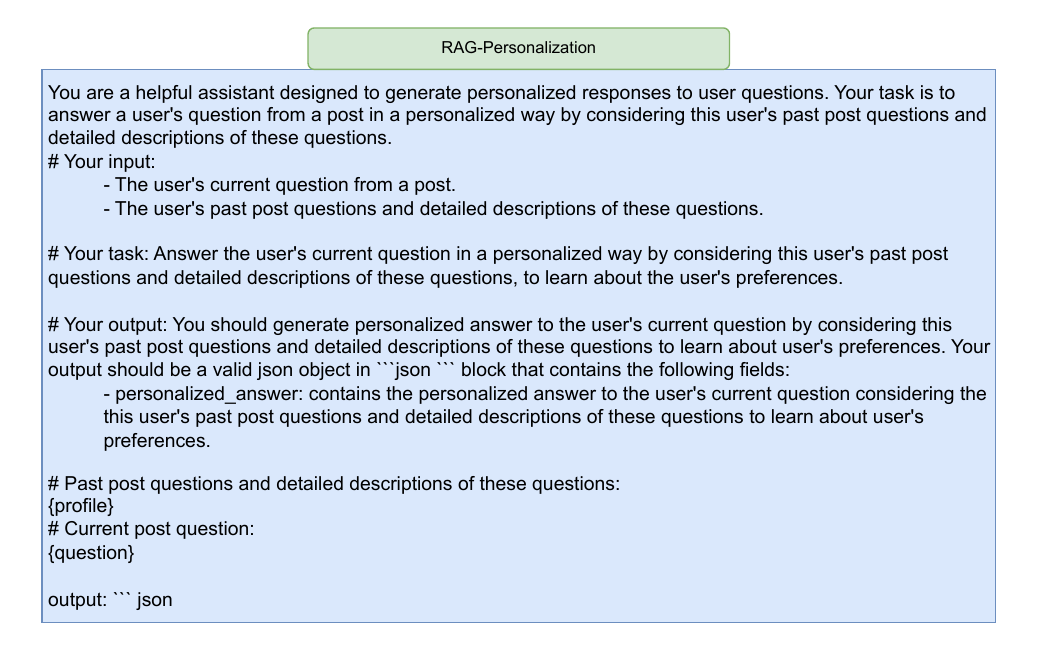}
    \caption{Prompt used with LLMs that personalize output using RAG.}
    \label{fig:baseline-rag-personalization}
\end{figure*}

\begin{figure*}[!ht]
    \centering \includegraphics[width=\textwidth]{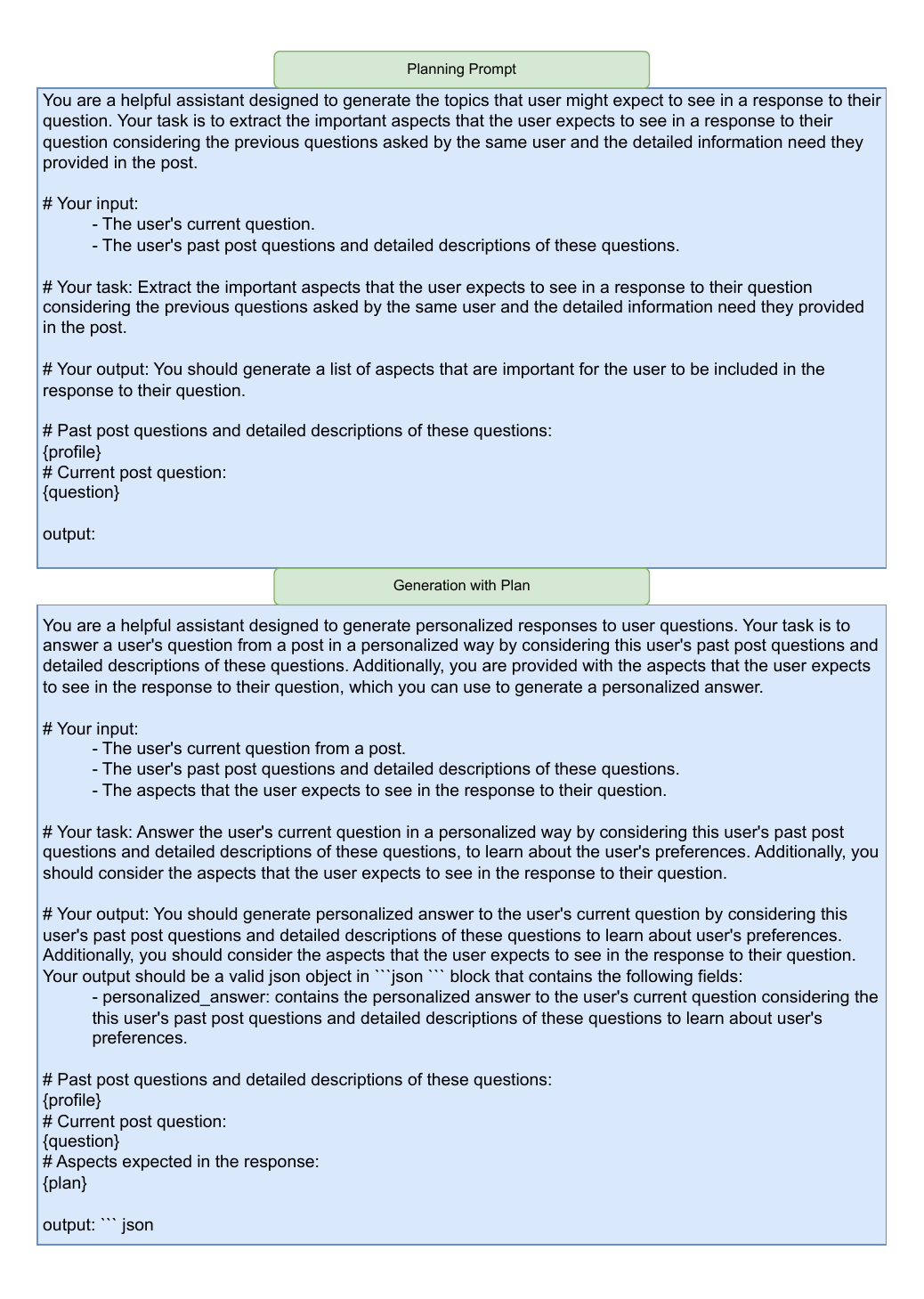}
    \caption{Prompt used with LLMs that personalize output using \ourmethodshort.}
    \label{fig:baseline-ours}
\end{figure*}

As described in Section~\ref{sec:method}, this paper proposes three baseline approaches for the \dataset benchmark. The first baseline, which generates responses without incorporating any personalized context, uses the prompt provided in Figure~\ref{fig:baseline-no-personalization}. The second baseline leverages RAG to incorporate personalized information retrieved from the user profile; it uses the prompt illustrated in Figure~\ref{fig:baseline-rag-personalization}. The third baseline further extends RAG by introducing an intermediate planning step that identifies the aspects the user expects in a response. These aspects are then used to guide the final response generation, following the prompt in Figure~\ref{fig:baseline-ours}.

\begin{table*}[!ht]
    \centering
    \adjustbox{max width=\textwidth}{\begin{tabular}{l|ccc|c}
    \toprule
    \multirow{2}{*}{\textbf{Method}} & \textbf{Arts \&} & \textbf{Lifestyle \& Personal} & \textbf{Society \&} & \textbf{Average}  \\
    & \textbf{Entertainment} & \textbf{Development} & \textbf{Culture} & \textbf{(macro)} \\
    \midrule
    \multicolumn{5}{c}{Gemma 2 Instruct (9B)} \\
    \midrule
    No Personalization  & 0.2025 & 0.3874 & 0.3973 & 0.3290 \\
    RAG-Personalization (Random $P$) & 0.1960 & 0.3330 & 0.3340 & 0.2876 \\
    RAG-Personalization (Asker's $P_u$) & 0.3260 & 0.4569 & 0.4765 & 0.4198 \\
    \midrule
    \ourmethodshort  & \textbf{0.3768$^\dagger$} & \textbf{0.4857$^\dagger$} & \textbf{0.5408$^\dagger$} & \textbf{0.4677$^\dagger$} \\
    \midrule
    \multicolumn{5}{c}{Qwen 2.5 Instruct (7B)} \\
    \midrule
    No Personalization  & 0.3419 & 0.4687 & 0.4566 & 0.4224 \\
    RAG-Personalization (Random $P$) & 0.2547 & 0.3789 & 0.3829 & 0.3388 \\
    RAG-Personalization (Asker's $P_u$) & 0.3822 & 0.4679 & 0.4909 & 0.4470 \\
    \midrule
    \ourmethodshort  & \textbf{0.3890} & \textbf{0.5051$^\dagger$} & \textbf{0.5181$^\dagger$} & \textbf{0.4707$^\dagger$} \\
    \midrule
    \multicolumn{5}{c}{GPT 4o-mini} \\
    \midrule
    No Personalization  & 0.3923 & 0.5175 & 0.5072 & 0.4723 \\
    RAG-Personalization (Random $P$) & 0.3016 & 0.4205 & 0.4044 & 0.3743\\
    RAG-Personalization (Asker's $P_u$) & 0.4357 & 0.4960 & 0.5179 & 0.4832 \\
    \midrule
    \ourmethodshort  & \textbf{0.4789$^\dagger$} & \textbf{0.5684$^\dagger$} & \textbf{0.5885$^\dagger$} & \textbf{0.5452$^\dagger$} \\
    \bottomrule
    \end{tabular}}
    \caption{Performance on the validation set. $^\dagger$ shows a statistically significant difference between the best-performing baseline and the others using t-test ($p < 0.05$).}
    \label{tab:main-results-dev}
\end{table*}

\begin{figure*}[!ht]
    \centering
    \includegraphics[width=\textwidth]{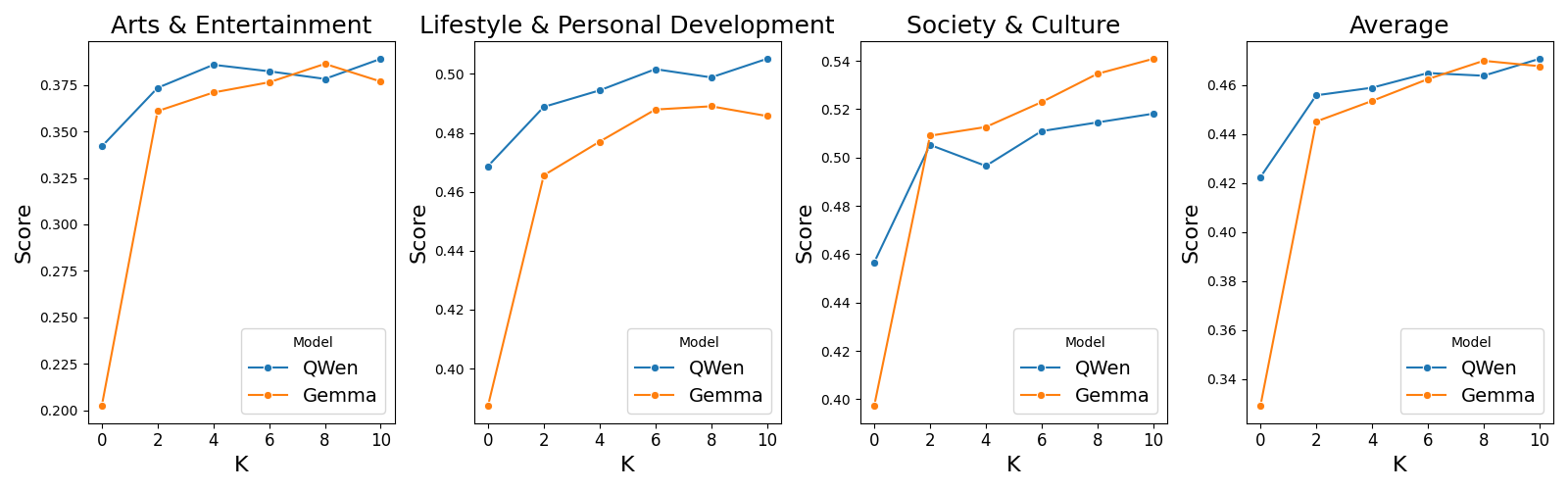}
    \caption{Effect of number of retrieved items from the user's profile on the performance of \ourmethod on the validation set.}
    \label{fig:ctx-size-perf-dev}
\end{figure*}

\section{Experimental Setup Details}
\label{app:setup}

In this paper, we use a combination of open and proprietary models for the generator LLM \( M \). Specifically, we employ instruction-tuned Gemma 2 (9B parameters\footnote{Available at: \url{https://hf.co/google/gemma-2-9b-it}}) \cite{gemma2}, instruction-tuned Qwen 2.5 (7B parameters\footnote{Available at: \url{https://hf.co/Qwen/Qwen2.5-7B-Instruct}}) \cite{qwen2.5}, and GPT-4o-mini\footnote{Available at: \url{https://openai.com/index/gpt-4o-mini-advancing-cost-efficient-intelligence/}} \cite{gpt4o} as the proprietary model. Throughout all experiments, the generator model remains frozen and is not fine-tuned, allowing us to isolate the effects of personalization methods without altering the underlying LLM. For the planner model \( M_{\text{plan}} \), we use Qwen 2.5 with 7B parameters. Training is performed using the Adam optimizer \citep{adam} with a learning rate of \( 5 \times 10^{-5} \) and a batch size of 32. We apply gradient clipping with a maximum norm of 1 to ensure stability. Training is conducted for up to 2000 steps, with a warmup phase spanning the first 10\% of steps, followed by a linear decay of the learning rate. We fine-tune the model using LoRA \cite{lora} with rank \( r = 16 \), scaling factor \( \alpha = 16 \), and a dropout rate of $0.05$, applied without modifying bias parameters. LoRA is implemented via the PEFT library.\footnote{Available at: \url{https://github.com/huggingface/peft}} Model checkpoints are evaluated every 250 steps using the validation set to monitor performance and select the best checkpoint.

All experiments are conducted using 4 NVIDIA A100 GPUs with 80GB of VRAM and 128GB of system RAM. All models are configured with a maximum input-output token limit of 8192 tokens. Response generation is performed using nucleus sampling \cite{nu_sampling} with a temperature of 0.1. For efficient inference and deployment of LLMs, we leverage the VLLM library.\footnote{Available at: \url{https://github.com/vllm-project/vllm}} For the retriever, we use Contriever \cite{contriever}, a dense retrieval model fine-tuned on the MS MARCO dataset \cite{msmarco}, to retrieve \( k = 10 \) relevant items from the user profile \( P_u \), unless otherwise specified. 

\section{Experiments on validation set}
\label{app:dev-perf}

The results of the baselines on the validation set of the \dataset benchmark are reported in Table~\ref{tab:main-results-dev}. These results confirm the findings discussed in Section~\ref{sec:main-results} on the test set, demonstrating consistent trends across both the test and validation sets. Additionally, the effect of varying the number of retrieved items from the user profile on the performance of \ourmethod is shown in Figure~\ref{fig:ctx-size-perf-dev}. These results mirror the findings discussed in Section~\ref{sec:main-results} for the test set, reinforcing that incorporating more personalized context from the user profile leads to improved performance on the validation set.

\section{AI Assistance Usage}

We used ChatGPT\footnote{\url{https://chat.openai.com/}} as a writing assistant. Specifically, initial drafts of certain paragraphs were paraphrased using ChatGPT, after which manual revisions were applied before inclusion in the paper.

\end{document}